\documentclass{article}
\usepackage{iclr2027_conference,times}
\usepackage{amsmath,amsfonts,bm}

\def\eqref#1{equation~\ref{#1}}
\def\1{\bm{1}}

\DeclareMathAlphabet{\mathsfit}{\encodingdefault}{\sfdefault}{m}{sl}
\SetMathAlphabet{\mathsfit}{bold}{\encodingdefault}{\sfdefault}{bx}{n}

\usepackage{amssymb}
\usepackage{graphicx}
\usepackage{subcaption}
\usepackage{booktabs}
\usepackage{xcolor}
\usepackage{hyperref}
\usepackage{url}
\usepackage{multirow}
\usepackage{wrapfig}
\usepackage{arydshln}
\usepackage{tikz}

\definecolor{mjRevision}{HTML}{B2186B}
\definecolor{mjComment}{HTML}{007A78}
\definecolor{shjhRevision}{HTML}{B85C00}
\definecolor{shjhComment}{HTML}{7040A0}

\definecolor{maskRed}{HTML}{B3261E}
\definecolor{maskOrange}{HTML}{B85C00}
\definecolor{maskYellow}{HTML}{997300}
\definecolor{maskGreen}{HTML}{287D3C}
\definecolor{maskBlue}{HTML}{1565C0}
\definecolor{maskIndigo}{HTML}{403C8F}
\definecolor{maskViolet}{HTML}{7B3294}
\definecolor{maskPink}{HTML}{A12663}
\newcommand{\method}{AgentDreamer}

\newcommand{\ourmetric}{Motion-IG}

\title{\method{}: Agentic 3D Motion Reasoning for Text-Controllable Physics Simulation}
\author{
Mijeong Kim$^{*}$ \hspace{1.5em}
Suho Park$^{*}$ \hspace{1.5em}
Junha Park \hspace{1.5em}
Bohyung Han\\[0.5em]
{\normalfont\small $^{*}$Equal contribution}
}
\iclrfinalcopy
\begin{document}
\maketitle
\lhead{\small\method{}: Agentic 3D Motion Reasoning for Text-Controllable Physics Simulation}

% !TEX root = ../main_agentdreamer.tex
\begin{abstract}
%We propose \method{}, a framework for text-controllable physics simulation that uses an agent to connect the text and motion by providing motion planning. 
%Given a text instruction and scene geometry, the agent reasons about the requested physical behavior and expresses it through parametric functions over temporal chunks.
%This formulation draws on the language understanding and physical knowledge of the agent to produce dense trajectories %The resulting trajectories provide pseudo supervision that connects language descriptions to physical motion.
%We optimize a differentiable physics simulator to match these trajectories, translating the motion plan into simulated dynamics.
%We further introduce \ourmetric{}, an information-gain-based metric for text-motion alignment through masked text recovery.
%The metric measures how much simulated motion improves the recovery of masked motion descriptions compared with recovery without the motion.
%Experimental comparisons demonstrate that agent-generated motion priors improve the alignment of simulated motion with text instructions, better reproducing physical behaviors that video-based guidance can conflate.
%
We propose AgentDreamer, an agent-guided optimization framework that aligns text instructions with simulated motion through explicit 3D motion planning.
Given a text instruction and scene geometry, an off-the-shelf agent translates the instruction into a structured semantic motion prior, specifying what should move, in which direction, and in what temporal order.
The agent represents this prior using parametric functions over a series of temporal chunks, yielding dense 3D trajectories.
We then optimize the parameters of a differentiable MPM simulator such that its rollout best matches the prior while remaining constrained by the scene dynamics.
This allows the agent to provide semantic guidance from language, while the simulator grounds it in physically feasible motion.
We further introduce Motion-IG, an information-theoretic metric for text--motion alignment based on masked text recovery.
By comparing the likelihood of recovering masked motion-related content with and without the simulated motion, Motion-IG measures the information contributed by the motion beyond that available from the remaining text and scene context.
Experiments show that AgentDreamer improves text alignment over video-based guidance, producing simulations that more closely follow the requested physical behavior.
Our project page is available at \url{https://cv.snu.ac.kr/research/AgentDreamer}.

\iffalse
We propose \method{}, a framework for text-controllable physics simulation that bridges text and simulated motion through agent-based 3D motion planning.
We formulate motion planning as a physical reasoning problem, drawing on the language understanding and physical knowledge of the agent.
Given a text instruction and scene geometry, the agent infers the requested motion through parametric functions over temporal chunks.
The resulting trajectories provide pseudo supervision for optimizing a differentiable physics simulator to produce physically plausible motion aligned with the text.
We further introduce \ourmetric{}, an information-gain-based metric that measures text-motion alignment through masked text recovery.
By comparing recovery likelihoods with and without simulated motion, the metric measures how much the motion contributes to recovering the masked content beyond the remaining text.
Experiments demonstrate that \method{} improves text alignment over video-based guidance, producing simulations that more closely follow the requested physical behavior.
\fi
\end{abstract}
% !TEX root = ../main_agentdreamer.tex
\section{Introduction}
\label{sec:intro}

Controlling physical motion in simulated scenes is important for creating dynamic environments for embodied AI and interactive content generation.
Without an intuitive control interface, users must manually tune simulation parameters through repeated simulation and adjustment to obtain the desired motion.
Text-controllable physics simulation offers a promising alternative by allowing users to directly describe the intended behavior in natural language.
Recent approaches, including PhysFlow~\citep{physflow} and MotionPhysics~\citep{wang2026motionphysics}, leverage video generative priors to guide the optimization of simulation parameters.
In this way, they connect high-level user specifications to the dynamics of a 3D scene.

\iffalse
Text-controllable physics simulation is a promising direction for creating dynamic environments for embodied AI and interactive content generation.
Without text-based control, users must manually tune simulation parameters to obtain the desired motion, often through repeated simulation and adjustment.
Natural-language instructions provide a more accessible interface by allowing users to describe the intended behavior directly.
Recent approaches, including PhysFlow~\citep{physflow} and MotionPhysics~\citep{wang2026motionphysics}, use video generative priors to translate these descriptions into motion guidance.
The generated motion serves as a reference for optimizing simulation parameters, connecting language input to the dynamics of a 3D scene.
\fi

Despite this progress, the resulting simulation may not reproduce the physical behavior described by the text.
A central difficulty lies in the motion reference itself: visually plausible video generation does not ensure that the requested dynamics are represented correctly.
For example, a request for a rolling ball may produce sliding, omitting the rotation that distinguishes the two behaviors.
Optimizing a simulator to follow such a reference can therefore yield physically plausible motion that remains inconsistent with the text.
This discrepancy motivates explicit 3D motion planning that describes how the object should move to reproduce the requested behavior.

\begin{figure}[t]
    \centering
    \includegraphics[
        width=0.99\linewidth,
    ]{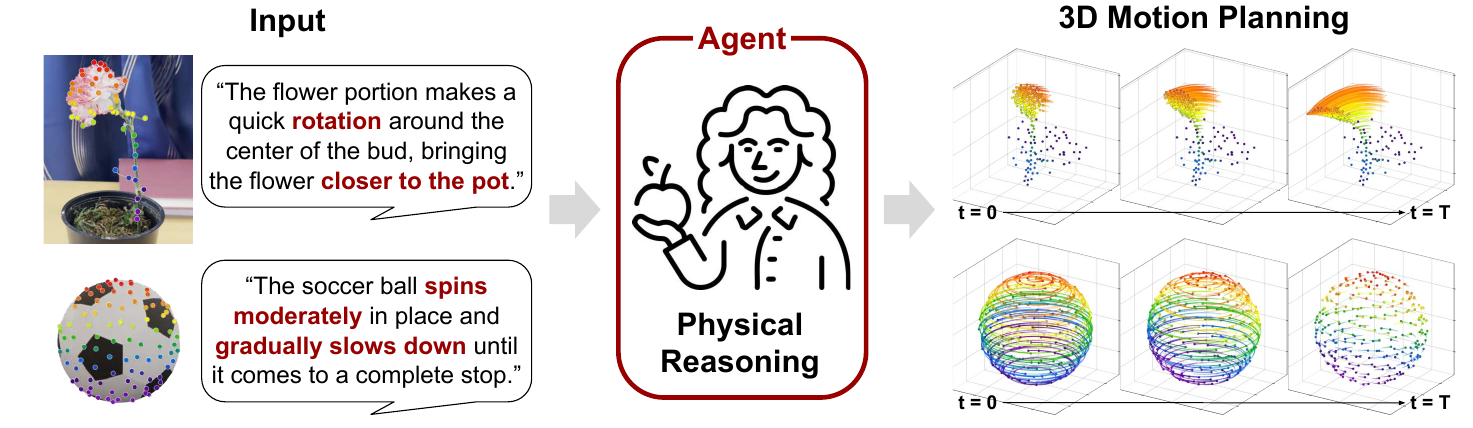}
    \vspace{-2mm}
    \caption{
    3D motion planning as physical reasoning.
    Given a text instruction and scene geometry, the agent reasons about the physical behavior required by the instruction and constructs a corresponding parametric motion plan.
    The plan produces dense 3D trajectories that guide physics optimization toward the requested motion.
    }
    \label{fig:agent_generated_text2motion}
    \vspace{-2mm}
\end{figure}

We propose \method{}, a framework for text-controllable physics simulation that bridges text and simulated motion through agent-based 3D motion planning.
Our key idea is to formulate motion planning as a physical reasoning problem.
Given a text instruction and scene geometry, an off-the-shelf multimodal agent draws on its language understanding and physical knowledge to infer how the object should move, as shown in Figure~\ref{fig:agent_generated_text2motion}.
The agent thus takes a direct role in constructing the motion plan, extending its use beyond estimating material parameters~\citep{physflow, wang2026motionphysics}. 
This plan provides explicit 3D guidance for translating the text into simulated motion.

To support motion reasoning, we introduce automatic semantic injection that associates sampled points with visible object parts.
We provide a scene image with colored point markers corresponding to the 3D point input, together with coordinate-system information.
This correspondence helps the agent interpret the role of each point in the requested motion.
The agent then divides the time horizon into temporal chunks and infers a parametric motion function for each.
These functions describe distinct motion stages and provide dense trajectories without predicting each future position independently.
The resulting trajectories serve as pseudo supervision for a differentiable Material Point Method (MPM) simulator.
We optimize simulation parameters to match these trajectories, guiding the simulator toward physically plausible motion aligned with the text.

We further introduce \ourmetric{}, an information-gain-based metric for measuring text-motion alignment through masked text recovery.
An MLLM-as-a-judge score~\citep{zheng2023judging,chen2024mllm}, read as the model's verbalized confidence, considers the complete text and simulated motion together, without isolating the information provided by the motion.
We instead mask motion-related content in the text and compare its recovery likelihoods with and without simulated motion under a fixed evaluation model.
The difference measures how much the motion contributes to recovering the masked content beyond the remaining text.

Experiments demonstrate that \method{} improves text alignment over video-based guidance, producing simulations that more closely follow the requested physical behavior.
Our contributions are summarized as follows:

\begin{itemize}
    \item We propose \method{}, a framework for text-controllable physics simulation that bridges text and simulated motion through agent-based 3D motion planning.
    \item We formulate motion planning as physical reasoning through parametric functions over temporal chunks, providing dense trajectories as pseudo supervision for differentiable physics optimization.
    \item We introduce \ourmetric{}, an information-gain-based metric for text-motion alignment that compares masked text recovery likelihoods with and without simulated motion.
\end{itemize}

\section{Related Works}
\label{sec:related}

\paragraph{Dynamic 3D representations}
Recent methods extend 3D Gaussian Splatting (3DGS)~\citep{kerbl20233d} to reconstruct dynamic scenes from video sequences.
These methods~\citep{luiten2024dynamic3dgaus, wu20244dgs, yang2024deformable, lin2024gaussianflow, katsumata2024compact, jiang2025timeformer, lee2024fully} represent temporal changes through persistent Gaussian trajectories, deformation fields, or space-time primitives.
The particle-based structure of Gaussian primitives further enables these representations to be extended to physics-based dynamics.
PhysGaussian~\citep{xie2024physgaussian} couples Gaussian primitives with Material Point Method (MPM) simulation, using a shared representation for physical dynamics and rendering.
Subsequent works develop elastic deformation models~\citep{springgau}, estimate material properties~\citep{zhang2024physdreamer,cai2024gic}, and support diverse material behaviors and interactions~\citep{lin2025omniphysgs,jiang2025phystwin}.

\iffalse
Physics-based motion generation incorporates physical principles to model realistic object dynamics and interactions.
Differentiable simulation frameworks, such as differentiable Material Point Method (MPM)~\citep{mpm}, enable gradient-based optimization of physical parameters and motion while capturing complex material behaviors.
Early works~\citep{li2023pac} couple MPM with Neural Radiance Fields (NeRFs)~\citep{mildenhall2021nerf} to integrate physics simulation with learned 3D representations, but the continuous implicit nature of NeRFs makes it non-trivial to establish correspondence with discrete simulation particles.
With the emergence of 3D Gaussian Splatting (3DGS)~\citep{kerbl20233d}, PhysGaussian~\citep{xie2024physgaussian} demonstrates that explicit Gaussian primitives can be naturally coupled with particle-based physics simulation.
Subsequent works extend this paradigm to model diverse deformations~\citep{springgau}, material properties~\citep{zhang2024physdreamer, cai2024gic}, and physical interactions~\citep{jiang2025phystwin, lin2025omniphysgs}.
\fi

\vspace{-2mm}
\paragraph{Text-conditional physics simulation}
Text provides an intuitive interface for specifying object dynamics and interactions.
Recent approaches~\citep{wang2026motionphysics, physflow,liu2024physics3d, lin2025omniphysgs} translate these instructions into simulation guidance through video diffusion priors.
PhysFlow~\citep{physflow} uses optical-flow guidance to optimize physical parameters, while MotionPhysics~\citep{wang2026motionphysics} employs learnable motion distillation to guide simulation toward text-specified dynamics.
These approaches primarily derive motion guidance in 2D space from generated videos. In contrast, our method constructs an explicit 3D motion prior through physical reasoning with an off-the-shelf multimodal agent, directly specifying the desired scene dynamics. Prior works have also used multimodal models to construct explicit motion priors, including bounding-box plans in VLIPP~\citep{yang2025vlipp} and 2D point trajectories in C-Drag~\citep{cdrag}. However, these representations remain sparse and confined to image space. Our approach instead produces dense 3D trajectories over the scene geometry, which serve as pseudo supervision for differentiable physics simulation.

\vspace{-2mm}
\paragraph{Agent-guided physics-aware motion evaluation}
Vision-language models provide semantic and physical feedback for evaluating and improving generated motion~\citep{lin2024vqascore, bansal2025videophy, meng2024towards}.
PhyGDPO~\citep{cai2026phygdpo} uses VLM-based scores of semantic adherence and physical commonsense to guide groupwise preference optimization.
PhyCo~\citep{narayanan2026phyco} evaluates generated videos through targeted physical questions and converts response logits into differentiable rewards for refining video generation.
These approaches assess motion through semantic or physical judgments~\citep{zheng2023judging,chen2024mllm}, but do not explicitly separate the evidence supplied by motion from information available in the textual context alone.
We instead formulate text-motion alignment as conditional information gain~\citep{wyner1978definition}, measuring how much a simulated trajectory improves the recovery of masked motion semantics over the remaining text.
%This provides an evaluator-based measure of the contribution of generated motion to interpreting the instruction.
% !TEX root = ../main_agentdreamer.tex
\section{Method}
\label{sec:method}

\begin{figure}[t]
    \centering
    \begin{subfigure}[b]{0.99\linewidth}
        \centering
        \includegraphics[width=\linewidth]{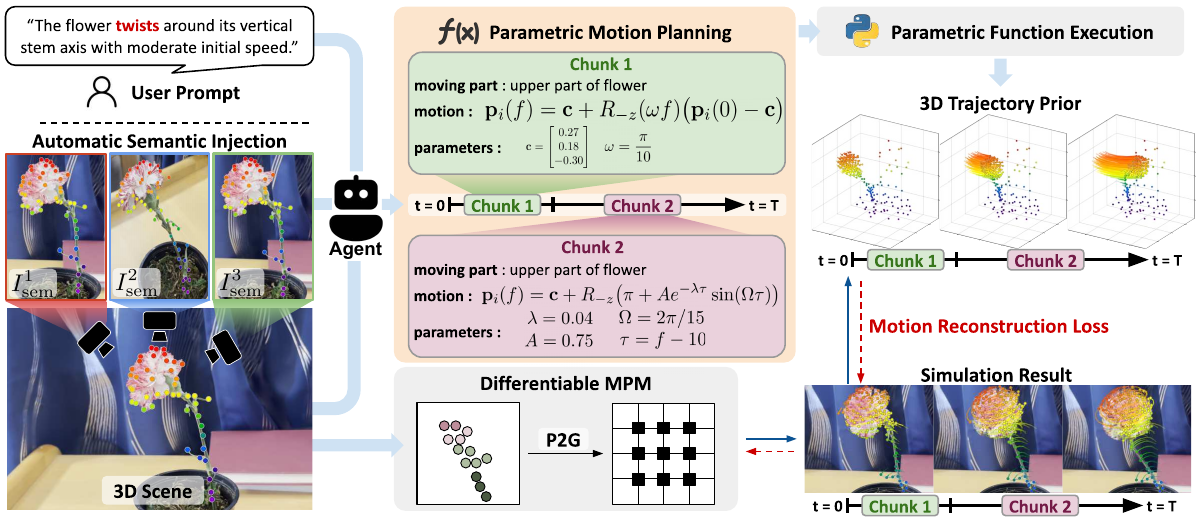}
    \end{subfigure}
    \vspace{-3mm}
    \caption{
    Overview of \method{}.
    Given a text instruction and scene geometry, an off-the-shelf agent constructs a structured semantic motion prior using parametric functions over temporal chunks.
    The resulting dense 3D trajectories guide optimization of a differentiable MPM simulator under the scene dynamics.
    }
    \label{fig:framework}
    \vspace{-3mm}
\end{figure}

We present \method{}, a framework for text-controllable physics simulation through agent-generated 3D motion planning, as shown in Figure~\ref{fig:framework}.
The agent translates the requested behavior into parametric motion functions, whose trajectories provide pseudo supervision for physics optimization.
We further introduce \ourmetric{} to measure text--motion alignment through masked text recovery.

\subsection{Agent-Guided Physics Simulation}
\label{subsec:physics_simulation}

Given a text instruction $X$ and initial scene geometry $\mathcal{G}$, we construct a 3D motion plan and optimize simulation parameters to follow its trajectories.

\subsubsection{Agent-Based Motion Planning}
\label{subsubsec:motion_planning}

\begin{figure}[t]
    \centering
    \includegraphics[width=\linewidth]{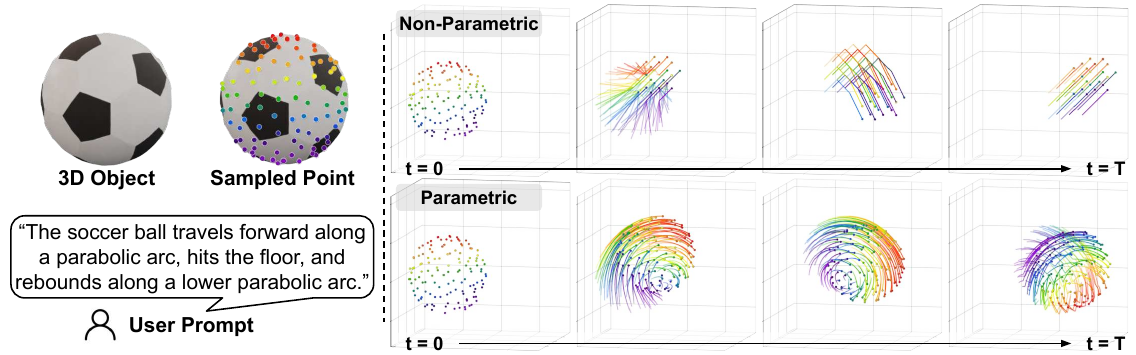}
    \caption{
    Effect of parametric motion planning.
    Direct prediction of future point positions produces irregular trajectories and distorts the object structure (top).
    Parametric functions yield coherent trajectories that better preserve spatial relationships between points (bottom).
    Colors identify corresponding points across time.
    }
    \label{fig:parametric_motion}
\end{figure}

\paragraph{Automatic semantic injection}
Point coordinates alone provide limited semantic information about the object parts relevant to the instruction $X$.
Let $\mathbf{p}_k^0$ be the initial position of MPM particle $k$, and let $\mathcal{K}$ be the set of particle indices selected for guidance.
The set $\mathcal{K}$ can contain any selected MPM particles; in our experiments, it consists of sampled particles.
For each initial view $I^v$, we project the selected points with camera projection $\pi^v$ and mark their indices as follows:
\begin{equation}
    I_{\mathrm{sem}}^v
    =\operatorname{Annotate}\!\left(I^v,
    \{(k,\pi^v(\mathbf{p}_k^0)):k\in\mathcal{K}\}\right).
    \label{eq:semantic_injection}
\end{equation}
Let $\mathbf{I}_{\mathrm{sem}}=(I_{\mathrm{sem}}^1,\ldots,I_{\mathrm{sem}}^V)$ denote the annotated views, where $V$ is the number of views.
Each marker displays the particle index $k$ consistently across views, allowing the agent to associate its 3D position with a visible object part.
We also specify the coordinate origin, axis directions, and spatial units.
Examples of annotated inputs are given in the Appendix.

\paragraph{Parametric motion planning}
Direct prediction of future point positions can distort the point configuration over time, as shown in Figure~\ref{fig:parametric_motion}.
We instead ask the agent to infer motion functions and their parameters.
Since a single function may not capture distinct motion stages, the agent divides the time horizon into temporal chunks and specifies a function for each.
The agent groups points with similar motion and divides the time horizon into $J$ chunks with boundaries $0=t_0<\cdots<t_J=T$.
Given the instruction, selected point coordinates $\mathcal{C}=\{(k,\mathbf{p}_k^0):k\in\mathcal{K}\}$, and annotated views $\mathbf{I}_{\mathrm{sem}}$, the agent $\mathcal{A}$ produces the plan as follows:
\begin{equation}
    \Pi=\mathcal{A}(X,\mathcal{C},\mathbf{I}_{\mathrm{sem}})
    =\{(\phi_j,t_{j-1},t_j)\}_{j=1}^{J}.
    \label{eq:agent_motion_plan}
\end{equation}
For chunk $j$, $\phi_j$ combines a center trajectory with the local motion of each point group.
The function $\phi_j(k,\mathbf{x},\tau)$ gives the target 3D position for point $k$ from its position $\mathbf{x}$ at the chunk start after elapsed time $\tau$.
Starting from $\widehat{\mathbf{p}}_k(0)=\mathbf{p}_k^0$, the target trajectory is given by
\begin{equation}
    \widehat{\mathbf{p}}_k(t)
    =\phi_j\!\left(k,\widehat{\mathbf{p}}_k(t_{j-1}),t-t_{j-1}\right),
    \qquad t\in[t_{j-1},t_j].
    \label{eq:parametric_motion}
\end{equation}
The condition $\phi_j(k,\mathbf{x},0)=\mathbf{x}$ ensures continuity between chunks.
Evaluating these functions at the desired times yields dense 3D trajectories.

\subsubsection{MPM Optimization}
\label{subsub:motion_prior}

\paragraph{MPM simulation}
The planned trajectories provide pseudo supervision for a differentiable Material Point Method (MPM) simulator.
We optimize the simulation parameters $\boldsymbol{\theta}$, including object-level material parameters and per-particle initial velocities, to reproduce the planned motion under the scene dynamics.
Let $\mathbf{s}^{n}$ denote the MPM state after simulation step $n$, including particle positions, velocities, and deformation-related quantities, and let $\Delta t_{\mathrm{sim}}$ be the step size.
The state evolves as
\begin{equation}
    \mathbf{s}^{n+1}
    =\operatorname{MPM}\!\left(\mathbf{s}^n;\boldsymbol{\theta},\Delta t_{\mathrm{sim}}\right).
    \label{eq:mpm_update}
\end{equation}
Let $\mathbf{p}_k(t;\boldsymbol{\theta})$ be the simulated position of particle $k$ at output time $t$.
The particle index $k$ establishes correspondence with the target position $\widehat{\mathbf{p}}_k(t)$.

\paragraph{Motion reconstruction loss}
We match the simulated and target positions of the selected particles at output times $t_1,\ldots,t_F$:
\begin{align}
    \mathcal{L}_{\mathrm{motion}}(\boldsymbol{\theta})
    &=
    \frac{1}{KF}
    \sum_{k\in\mathcal{K}}
    \sum_{f=1}^{F}
    \left\|
        \mathbf{p}_k(t_f;\boldsymbol{\theta})
        -
        \widehat{\mathbf{p}}_k(t_f)
    \right\|_2^2,
    \label{eq:motion_reconstruction}\\
    \boldsymbol{\theta}^{*}
    &=
    \arg\min_{\boldsymbol{\theta}}
    \mathcal{L}_{\mathrm{motion}}(\boldsymbol{\theta}).
    \label{eq:physics_optimization}
\end{align}
Here $K=|\mathcal{K}|$.
Although the loss is evaluated only for particles in $\mathcal{K}$, MPM evolves the full scene.
The motion functions provide target positions for the selected particles at each output time.
We keep the plan fixed and update $\boldsymbol{\theta}$ using gradients through the MPM rollout.

\subsection{Motion-IG}
\label{subsec:alignment_eval}

We introduce Motion-Conditioned Information Gain (\ourmetric{}) to measure text--motion alignment through masked text recovery.
It measures how much observing the simulated motion improves recovery of motion-related content beyond the remaining text.

\paragraph{Notation}
Let $X=(x_1,\ldots,x_L)$ be the instruction word sequence and $\mathcal{M}=\{m_1,\ldots,m_R\}$ the set of masked positions, where $m_1<\cdots<m_R$.
Replacing these words with [MASK] yields the masked sequence $X_m$; the removed words form the ordered subsequence $X^{\mathcal{M}}=(x_{m_1},\ldots,x_{m_R})$.
We mask motion words and attributes while retaining object and part names (Table~\ref{table:mask_examples}).
Let $Y$ denote the simulated motion, represented by rendered frames and serialized 3D trajectories.
The masked text $X_m$ and annotated initial views $\mathbf{I}_{\mathrm{sem}}$ remain identical in both conditions.
For brevity, we omit the shared conditioning on $\mathbf{I}_{\mathrm{sem}}$ in the equations below.

\begin{table}[t]
\centering
\caption{
Examples of word-level masking for \ourmetric{}.
Bold words in the original instruction $X$ are replaced by [MASK] in the masked instruction $X_m$.
}
\label{table:mask_examples}
\vspace{-2mm}
\resizebox{\linewidth}{!}{%
    \setlength{\tabcolsep}{17pt}
    \begin{tabular}{ll}
        \toprule
        \textbf{Sequence} & \textbf{Instruction} \\
        \midrule
        $X$ & The \textbf{soft} fox \textbf{sways} from \textbf{side} to side while \textbf{remaining} \textbf{attached} at the rear. \\
        $X_m$ & The [MASK] fox [MASK] from [MASK] to side while [MASK] [MASK] at the rear. \\
        \midrule
        $X$ & The basketball \textbf{rolls} \textbf{forward} in a \textbf{straight} \textbf{line}, \textbf{rotating} \textbf{naturally} \textbf{without} \textbf{slipping}. \\
        $X_m$ & The basketball [MASK] [MASK] in a [MASK] [MASK], [MASK] [MASK] [MASK] [MASK]. \\
        \midrule
        $X$ & The \textbf{rigid} playdoh ball \textbf{falls} \textbf{straight} \textbf{down} and \textbf{bounces} \textbf{several} \textbf{times} without \textbf{deforming}. \\
        $X_m$ & The [MASK] playdoh ball [MASK] [MASK] [MASK] and [MASK] [MASK] [MASK] without [MASK]. \\
        \bottomrule
    \end{tabular}
}
\end{table}

\paragraph{Information gain}
The masked content may be partially inferred from the remaining text and annotated views.
We therefore measure the additional information provided by the motion through conditional mutual information as follows:
\begin{align}
    I(X^{\mathcal{M}};Y\mid X_m)
    &=
    \mathbb{E}\!\left[
        \log
        \frac{
            p(X^{\mathcal{M}}\mid X_m,Y)
        }{
            p(X^{\mathcal{M}}\mid X_m)
        }
    \right].
    \label{eq:conditional_mi}
\end{align}
Since these probabilities are unknown, we use a fixed vision-language model $p_\psi$ to compute a pointwise score as follows:
\begin{align}
    s_{\psi}(X,Y)
    &=
    \log
    \frac{
        p_{\psi}(X^{\mathcal{M}}\mid X_m,Y)
    }{
        p_{\psi}(X^{\mathcal{M}}\mid X_m)
    }.
    \label{eq:sample_information}
\end{align}
The denominator gives the recovery likelihood without the simulated motion.
A positive score indicates that observing the motion increases the likelihood of the removed words; a negative score indicates that it decreases this likelihood.

\paragraph{Score computation}
We provide the masked text as the query and teacher-force the original instruction as the response, with and without simulated motion.
Let $N$ be the number of evaluated text--motion pairs.
For pair $(X_r,Y_r)$, let $(z_{r,1},\ldots,z_{r,L_r})$ denote the response tokens.
The token log-probability gain is given by
\begin{align}
    \delta_{\psi}(r,\ell)
    &=
    \log p_{\psi}
    \left(z_{r,\ell}\mid X_{m,r},Y_r,z_{r,<\ell}\right)
    -
    \log p_{\psi}
    \left(z_{r,\ell}\mid X_{m,r},z_{r,<\ell}\right).
    \label{eq:token_information_gain}
\end{align}
The preceding response tokens $z_{r,<\ell}$ are identical in both conditions.
Let $\mathcal{M}_r^{\mathrm{score}}\subseteq\mathcal{M}_r$ contain the masked positions selected for scoring, excluding words such as ``is'' and ``the'' but retaining negations such as ``not''.
For each selected word at position $m$, we score its first subword token at position $\ell_r(m)$.
Averaging first over words and then over $N$ instructions gives \ourmetric{} as follows:
\begin{align}
    \text{Motion-IG}
    &:=
    \frac{1}{N}
    \sum_{r=1}^{N}
    \frac{1}{|\mathcal{M}_r^{\mathrm{score}}|}
    \sum_{m\in\mathcal{M}_r^{\mathrm{score}}}
    \delta_{\psi}\bigl(r,\ell_r(m)\bigr).
    \label{eq:motion_ig}
\end{align}
This first-subword score is a proxy for Eq.~\ref{eq:conditional_mi}.
We compute \ourmetric{} for each method using its simulated motions while keeping the instructions, masks, annotated views, and evaluation models identical across methods.

% !TEX root = ../main_agentdreamer.tex
\section{Experiments}
\label{sec:experiments}

%We evaluate \method{} to examine whether agent-generated motion priors improve text-controllable physics simulation.
%We compare the resulting motion with existing methods, analyze the contributions of our design choices, and validate the proposed alignment metric against human judgments.

We evaluate whether agent-generated 3D motion plans improve text-controllable physics simulation across diverse object motions.
We use GPT-5.6 Sol as the off-the-shelf agent for 3D motion planning throughout our experiments.
We compare the simulations with video-guided methods using \ourmetric{} and blinded human rankings of text--motion alignment.

\subsection{Experimental Setup}

\paragraph{Scenes and instructions.}
We conduct experiments on eight scenes with 20 text instructions.
The dataset comprises four real-world scenes used by PhysFlow~\citep{physflow} (Alocasia, Carnation, Telephone and Fox) and four synthetic scenes, one from PAC-NeRF~\citep{li2023pac} (Playdoh) and three from GaussianVerse~\citep{xiang2025gaussianverse} (Basketball, Soccer Ball and Can).
The instructions cover falling and bouncing, rolling, slowing rotation, oscillation, and bending or deformation, with 10 applied to the real-world scenes and 10 to the synthetic scenes.

%Both methods additionally require externally specified forces, preventing their direct application under our input setting, where only the text instruction and scene geometry are provided.
%To accommodate this requirement, we use an agent to initialize their force inputs from the same instruction and scene context.
%\method{} requires no externally specified forces and directly guides simulation optimization with agent-generated 3D motion priors.

\vspace{-2mm}
\paragraph{Baselines and evaluation.}
We compare \method{} with PhysFlow~\citep{physflow} and MotionPhysics~\citep{wang2026motionphysics}, which use text-conditioned video diffusion to guide physics simulation.
We use four open-weight evaluators: InternVL3.5-8B~\citep{wang2025internvl3}, Qwen3-VL-8B~\citep{bai2025qwen3vl}, Qwen3.5-9B~\citep{qwen3.5}, and Gemma-4-12B~\citep{team2026gemma}.
We additionally report results with five API-based evaluators: GPT-5.6 Sol and GPT-5.6 Terra~\citep{openai2026gpt56}, GPT-6 Astra~\citep{openai2026gpt6astra}, Gemini 3.1 Pro~\citep{google2026gemini31pro}, and Gemini 3.8 Flash~\citep{google2026gemini38flash}.
To compute \ourmetric{}, we use teacher-forced token likelihoods for open-weight evaluators and directly requested probabilities for API-based evaluators.
For human evaluation, participants rank three anonymized simulations for each instruction by how well the motion follows the text.
%We evaluate all pairwise combinations of the compared methods for each instruction, with method identities hidden and presentation order randomized.

\iffalse
We report the trajectory-conditioned information score
$S_{\psi}^{(a)}$ defined in Section~\ref{subsec:alignment_eval}.
All methods share the same evaluator, masks, and trajectory representation.
We report the information score $S_{\psi}^{(a)}$ of Section~\ref{subsec:alignment_eval} in nats per masked word.
All methods share the same instructions, masks, initial-state image and evaluators; only the block describing the generated motion differs, and the condition without it is identical across methods.
We use four open vision-language models as evaluators, Qwen3-VL-8B, Qwen3.5-9B, Gemma-4-12B and InternVL3.5-8B, and report both the per-evaluator scores and their average, under both mask sets of Section~\ref{subsec:alignment_eval}.
We additionally conduct a human evaluation of text alignment and physical plausibility as separate criteria.
Participants view the instruction and rendered simulation and rate each criterion on a five-point scale.
Method identities are hidden, and presentation order is randomized.
\fi

\subsection{Comparison with Existing Methods}

\subsubsection{Qualitative Results}

\begin{figure}[t]
    \centering
    \includegraphics[width=\linewidth]{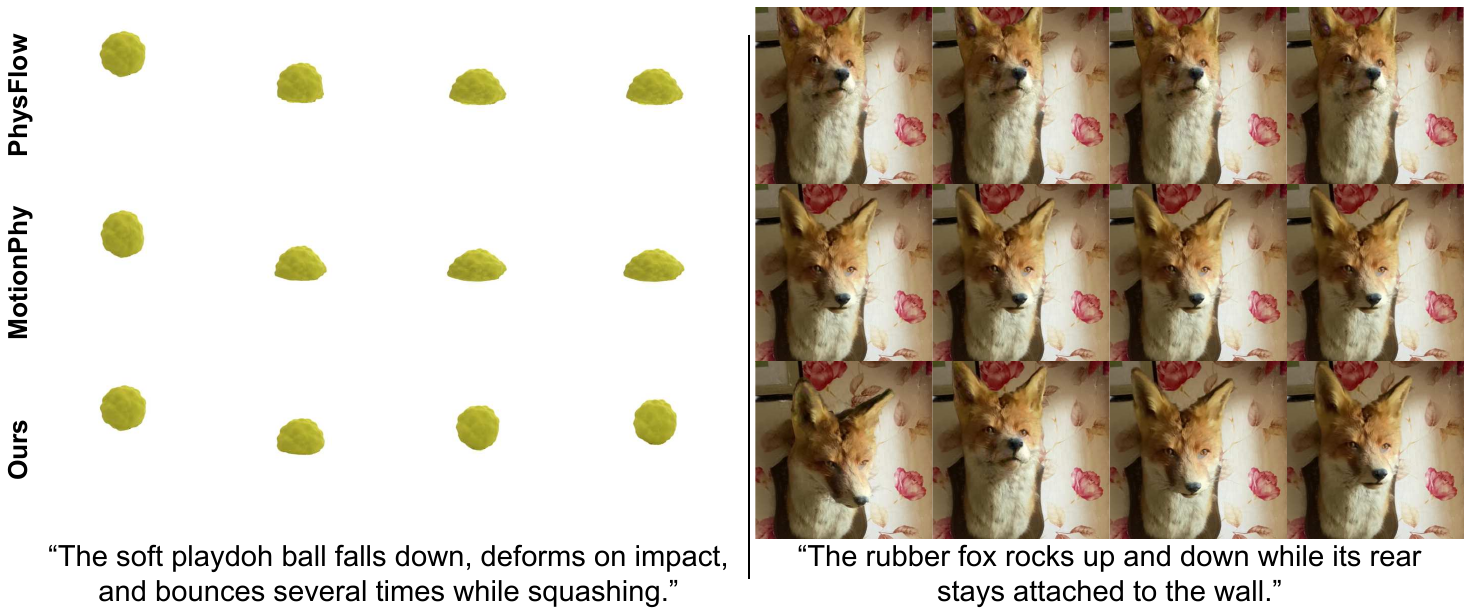}
    \caption{
    Qualitative comparison of text-controllable physics simulation.
    \method{} more closely follows the requested motion than PhysFlow and MotionPhysics in the shown examples.
    The shown examples contrast the requested motion with the behaviors produced by video-guided optimization and explicit 3D planning.
    }
    \label{fig:qualitative_results}
\end{figure}

Figure~\ref{fig:qualitative_results} compares the three methods on a bouncing playdoh ball and a fox instructed to rock while its rear remains attached.
For the ball, PhysFlow and MotionPhysics deform on impact but remain near the ground, whereas \method{} rebounds after compression as requested.
For the fox, the shown frames of \method{} display the requested rocking motion while the rear remains attached; the baseline sequences show less of this motion.
Both examples depend on the motion sequence specified in the instruction, which the agent represents through a 3D trajectory prior before simulation.

\subsubsection{Quantitative Results}

\begin{table}[t]
\centering
\caption{
Comparison of text--motion alignment using \ourmetric{} with
(a) token likelihoods from open-weight evaluators and
(b) verbalized probabilities from black-box API evaluators
(higher is better).
\textbf{Best in} counts the evaluator--instruction pairs in which each method ranks first.
}
\label{table:quantitative_results}
\vspace{-1.5mm}
\captionsetup[subtable]{position=top,font={small,bf},skip=1mm}

\begin{subtable}{\linewidth}
    \centering
    \caption{Open-weight evaluators}
    \label{table:main_open_weight}
    \scalebox{0.78}{%
        \setlength{\tabcolsep}{8.5pt}
        \begin{tabular}{l|cccc|c|c}
            \toprule
            \textbf{Method}
            & \textbf{Gemma-4}
            & \textbf{Qwen3.5}
            & \textbf{Qwen3-VL}
            & \textbf{InternVL3.5}
            & \textbf{Average}
            & \textbf{Best in} \\
            \midrule
            PhysFlow~\citep{physflow}
            & 0.36 & 0.62 & 0.58 & 0.52 & 0.52 & 15/80 \\
            MotionPhysics~\citep{wang2026motionphysics}
            & 0.13 & 0.58 & 0.81 & 0.56 & 0.52 & 22/80 \\
            \hdashline
            \method{} (ours)
            & \textbf{0.63} & \textbf{0.67}
            & \textbf{0.92} & \textbf{0.64}
            & \textbf{0.71} & \textbf{43/80} \\
            \bottomrule
        \end{tabular}
    }
\end{subtable}\par

\vspace{2mm}

\begin{subtable}{\linewidth}
    \centering
    \caption{Black-box evaluators}
    \label{table:main_black_box}
    \scalebox{0.78}{%
        \setlength{\tabcolsep}{8.5pt}
        \begin{tabular}{l|ccc|cc|c|c}
            \toprule
            & \multicolumn{3}{c|}{\textbf{GPT}}
            & \multicolumn{2}{c|}{\textbf{Gemini}}
            & & \\
            %\cmidrule(lr){2-4} \cmidrule(lr){5-6}
            \textbf{Method}
            & \textbf{6 Astra}
            & \textbf{5.6 Sol}
            & \textbf{5.6 Terra}
            & \textbf{3.8 Flash}
            & \textbf{3.1 Pro}
            & \textbf{Average}
            & \textbf{Best in} \\
            \midrule
            PhysFlow~\citep{physflow}
            & $-0.13$ & $-0.45$ & $+0.04$ & $-0.69$ & $+0.14$
            & $-0.22$ & 23/100 \\
            MotionPhysics~\citep{wang2026motionphysics}
            & $-0.49$ & $-0.96$ & $-0.44$ & $-0.93$ & $-0.32$
            & $-0.63$ & 15/100 \\
            \hdashline
            \method{} (ours)
            & $\mathbf{+0.56}$ & $\mathbf{+0.25}$ & $\mathbf{+0.48}$
            & $\mathbf{+0.12}$ & $\mathbf{+1.00}$
            & $\mathbf{+0.48}$ & \textbf{62/100} \\
            \bottomrule
        \end{tabular}
    }
\end{subtable}\par
\vspace{-3mm}
\end{table}

\paragraph{\ourmetric{} with open-weight evaluators}
Table~\ref{table:main_open_weight} compares text--motion alignment using four open-weight evaluators.
On the 20 evaluated instructions, \method{} achieves the highest score under all four evaluators.
Its positive scores indicate that providing the simulated motion increases the likelihood of the correct masked words, on average, compared with the condition without motion.
These results suggest that motion generated by \method{} provides more useful information for recovering the requested behavior than the compared methods.

\paragraph{\ourmetric{} with API-based evaluators}
Table~\ref{table:main_black_box} reports results using verbalized probabilities from five API-based evaluators.
The lower average scores may partly reflect stronger recovery from the masked text and annotated views alone: if an evaluator already predicts the masked words with high probability, observing motion provides less additional gain.
Nevertheless, \method{} achieves the highest score under all five evaluators and ranks first in $62/100$ evaluator--instruction pairs.
This consistent ranking suggests that its simulated motions provide more useful evidence for recovering the requested behavior across both evaluation approaches.
We compare rankings within each evaluator group because verbalized probabilities and teacher-forced token likelihoods have different scales.

\vspace{-2mm}
\paragraph{Human evaluation}
\begin{wrapfigure}{r}{0.44\linewidth}
    \centering
    \vspace{-5mm}
    \includegraphics[width=\linewidth]{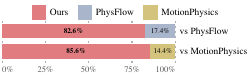}
    \caption{
    Pairwise human preference for text--motion alignment.
    Win rate is the percentage of trials in which one method is ranked above the other.
    \method{} is preferred over PhysFlow in 82.6\% of trials and over MotionPhysics in 85.6\% of trials.
    }
    \label{fig:human_pairwise}
\end{wrapfigure}
We conduct a human evaluation of text--motion alignment with 25 participants.
For each instruction, participants view the results of all three methods and rank them by how well they match the requested behavior.
\method{} is ranked first most often (73.6\%), compared with PhysFlow (14.0\%) and MotionPhysics (12.4\%).
Figure~\ref{fig:human_pairwise} further shows that \method{} is preferred over each baseline in pairwise comparisons.
This agrees with the \ourmetric{} ranking, providing further evidence that \method{} follows the text instructions more closely than the baselines.

\subsection{Analysis}

%\paragraph{Analysis of physical reasoning.}
\paragraph{Correspondence to classical mechanics.}

\begin{table}[t]
\centering
\caption{
Correspondence between agent-generated motion functions and expected forms across $20$ instructions.
The expected forms follow classical mechanics for rolling, spin decay, falling, and oscillation, while deformation uses smoothstep interpolation.
Expressions show representative components, and \textbf{Matched / Total} counts plans that use the expected form in each category.
}
\label{table:plan_functions}
\vspace{-1.5mm}
\scalebox{0.78}{%
    \setlength{\tabcolsep}{11pt}
    \begin{tabular}{lllc}
        \toprule
        \textbf{Motion cue}
        & \textbf{Function family}
        & \textbf{Representative expression}
        & \textbf{Matched / Total} \\
        \midrule
        Roll
        & Coupled translation and rotation
        & $d(\tau)=v\tau,\quad \theta(\tau)=(v/R)\tau$
        & 2/2 \\
        %\midrule
        Spin, slow down, stop
        & Quadratic
        & $\theta(\tau)=\omega_0\tau+\frac{1}{2}\alpha\tau^2,\quad \alpha<0$
        & 2/2 \\
        %\midrule
        Fall, bounce, jump
        & Quadratic
        & $z(\tau)=z_0+v_{z,0}\tau-\frac{1}{2}g\tau^2$
        & 4/4 \\
        %\midrule
        Sway, oscillate, rock
        & Sinusoid
        & $\theta(\tau)=A e^{-\lambda\tau}\sin(\omega\tau)$
        & 8/8 \\ %6/6 \\
        %Sway, oscillate, rock
        %& Sinusoid
        %& $\theta(\tau)=A\sin(\omega\tau)$
        %& 2/2 \\
        %\midrule
        Pull, bend, squash, deform
        & Smoothstep interpolation
        & $s(u)=3u^2-2u^3$
        & 4/4 \\
        \bottomrule
        \multicolumn{4}{l}{\small
        $\tau$: elapsed time within a chunk;
        $u=\tau/(t_j-t_{j-1})$: normalized time.} \\
        \multicolumn{4}{l}{\small
        $d$: travel distance;
        $\theta$: rotation angle;
        $R$: ball radius.} \\
    \end{tabular}
}
\vspace{-2mm}
\end{table}

We examine whether the agent-generated functions follow motion equations derived from classical mechanics.
Table~\ref{table:plan_functions} reports how often the plans match the expected function family.
The generated forms resemble the classical relations for rolling, falling, spin decay, and damped oscillation, including coupled translation and rotation, quadratic trajectories, and damped sinusoids.
For deformation, the agent uses smoothstep interpolation to describe the temporal change rather than a classical mechanics equation.
These similarities suggest that the agent draws on familiar physical motion patterns when constructing its 3D prior.

\vspace{-2mm}
\paragraph{Benefits of 3D motion planning across motion categories.}

\begin{figure}[t]
\centering
\begin{minipage}[t]{0.54\textwidth}
    \centering
    \includegraphics[width=\linewidth]{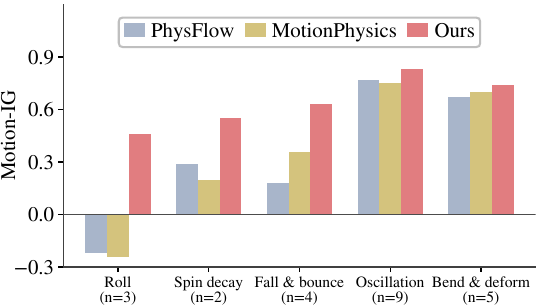}
\caption{
Text--motion alignment across motion categories measured by \ourmetric{} (higher is better).
\method{} achieves the highest score in every category; for rolling, both baselines have negative information gain.
}
    \label{fig:category_alignment}
\end{minipage}
\hfill
\begin{minipage}[t]{0.42\textwidth}
    \centering
    \includegraphics[width=\linewidth]{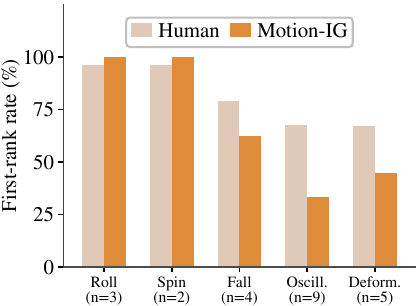}
\caption{
Consistency of \ourmetric{} with human evaluation across motion categories.
Both show higher first-rank rates for rolling and spinning and lower rates for oscillation and deformation.
}
    \label{fig:first_rank}
\end{minipage}
\vspace{-3mm}
\end{figure}

To examine the benefit of 3D planning over 2D video-based guidance, we compare \ourmetric{} across five motion categories in Figure~\ref{fig:category_alignment}.
\method{} achieves the highest score in every category, with the largest improvements for rolling, spin decay, and falling and bouncing.
Notably, both baselines obtain negative scores for rolling, indicating that their simulated motions make the masked behavior harder to recover than from the shared context without motion.
Rolling and sliding can appear similar in video, although rolling requires coupled translation and rotation in 3D.
The larger gain for rolling suggests that explicit 3D planning is especially useful when the requested motion is difficult to distinguish from video appearance alone.
Gains are smaller for oscillation and deformation, where the baselines already achieve relatively high scores.

\vspace{-2mm}
\paragraph{Consistency with human evaluation.}
We examine whether \ourmetric{} reflects human preferences by comparing how often \method{} ranks first among the evaluated methods in each motion category.
As shown in Figure~\ref{fig:first_rank}, both human evaluation and \ourmetric{} select our method most frequently for rolling and spinning.
The first-rank rates decrease for oscillation and deformation, where the baselines are selected more often.
Although the absolute rates differ, this shared trend suggests that \ourmetric{} captures category-dependent differences in text--motion alignment observed by human evaluators.
These category-level trends do not measure agreement on individual instruction rankings.

\vspace{-2mm}

\paragraph{Temporal visualization of motion planning}
\begin{figure}[t]
    \centering
    \includegraphics[width=\linewidth]{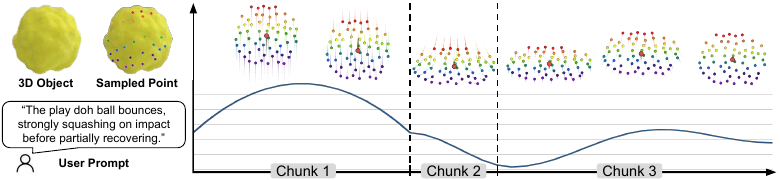}
    \caption{
    Temporal decomposition of an agent-generated motion plan.
    The agent uses separate parametric functions to describe the ball movement, compression on impact, and partial shape recovery.
    Dashed lines mark temporal chunks, and colored points show target positions over time.
    }
    \label{fig:chunk_visualization}
\end{figure}

Figure~\ref{fig:chunk_visualization} shows how the agent divides the requested motion into temporal chunks and constructs a parametric function for each.
The planned sequence captures the ball motion, strong compression on impact, and subsequent partial shape recovery.
The chunks correspond to these behavioral changes, suggesting that the agent identifies distinct motion stages within the instruction.
By describing each stage separately, the plan captures changes in physical motion that are difficult to represent with a single parametric function.

\vspace{-2mm}
\paragraph{Effect of planning components}%
\begin{figure}[t]
    \centering
    \begin{minipage}[t]{0.4\linewidth}
        \vspace{0pt}
        \centering
        \captionof{table}{
        Effect of planning components including semantic injection and parametric motion predictions.
        Parametric planning provides the largest individual gain, while combining both components achieves the highest score.
        }
        \label{table:component_ablation}
        \vspace{-2mm}
        \scalebox{0.70}{
            \setlength\tabcolsep{13pt}
            \begin{tabular}{cc|c}
                \toprule
                \textbf{Semantic} & \textbf{Parametric}
                & \multirow{2}{*}{\textbf{\ourmetric{}} $\uparrow$} \\
                \textbf{injection} & \textbf{planning} & \\
                \midrule
                & & 0.37 \\
                \checkmark & & 0.38 \\
                & \checkmark & 0.63 \\
                \hdashline
                \checkmark & \checkmark & \textbf{0.72} \\
                \bottomrule
            \end{tabular}
        }

    \end{minipage}
    \hfill
    \begin{minipage}[t]{0.57\linewidth}
        \vspace{0pt}
        \centering
        \captionsetup{skip=2pt}
        \includegraphics[width=\linewidth]{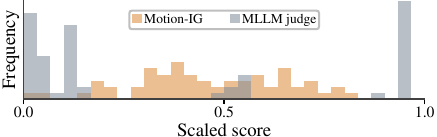}
        \vspace{-2mm}
        \caption{
        Score distributions on identical text--motion pairs, scored by gemma-4. 
        Motion-IG gives finer score resolution than an MLLM judge on the same text--motion pairs.
        }
        \label{fig:comparison_with_naive_evaluation}
    \end{minipage}
    \vspace{-2mm}
\end{figure}
Table~\ref{table:component_ablation} shows that both components improve text--motion alignment, with parametric planning providing the larger gain.
The small gain from semantic injection alone suggests that point annotations are most useful with parametric planning, where they help associate selected 3D points with visible object parts.
Without parametric planning, independently predicted point positions lose coherence over time (Figure~\ref{fig:parametric_motion}).
Motion functions instead provide coherent, dense target trajectories for physics optimization.

\vspace{-2mm}
\paragraph{Comparison with MLLM judge}
Figure~\ref{fig:comparison_with_naive_evaluation} compares \ourmetric{} with MLLM judgments~\citep{zheng2023judging,chen2024mllm} on the same $60$ text--motion pairs after scaling each score type to its own range.\footnote{The judge takes the full instruction and simulated motion as input and directly assigns a score in $[0,1]$ for how well the motion follows the instruction. We use its greedily decoded score as verbalized confidence.}
The direct judge scores the instruction and motion together, whereas \ourmetric{} measures the additional evidence that the motion provides for recovering masked motion words.
Although asked for a real-valued score, the MLLM judge assigns only $11$ distinct values to the $60$ pairs, giving the same score to multiple motions.
In contrast, \ourmetric{} assigns a distinct score to every pair in this comparison.
The reduced number of ties gives finer resolution when ranking these pairs; Figure~\ref{fig:first_rank} separately compares category-level first-rank rates with human evaluation.
%Evaluation prompts, scoring details, and additional results are provided in the supplementary material.

% !TEX root = ../main_agentdreamer.tex
\vspace{-2mm}
\section{Conclusion}
\vspace{-2mm}
We presented \method{}, a framework that uses an agent to generate 3D motion priors for text-controllable physics simulation.
Given a text instruction and scene geometry, the agent reasons about the underlying physics and constructs parametric functions to describe motion consistent with the instruction.
The resulting trajectories guide differentiable physics optimization to reproduce the requested motion.
We have also introduced \ourmetric{} to measure how much simulated motion helps recover masked motion words beyond the remaining text.
Experiments show improved text alignment over video-guided baselines, demonstrating the effectiveness of agentic motion priors.

\subsection*{AI use statement}
In this work, we used generative AI tools to help develop theoretical models or conceptual frameworks, design or provide feedback on research methodology or experiments, implement methods, and assist with translation.
We have not used generative AI tools to generate synthetic datasets, formulate mathematical claims, clean or reformat datasets, or interpret results, and tasks related to providing critical ingredients for proving mathematical claims, assisting in the writing of proofs, proposing or refining hypotheses, and supporting qualitative or thematic data analysis are not applicable to this work.

Additionally, we used generative AI tools to create or modify scientific figures or images, suggest experimental parameters, create or edit software code and research artifacts, draft parts of the manuscript, summarize or analyze existing literature, discover research topics or identify gaps, brainstorm research ideas, search for information, improve manuscript readability, identify relevant literature, format references, and suggest
manuscript structure.

We have reviewed all AI-assisted work. AI-assisted code was verified through execution and testing, and all methodological suggestions and translations were manually reviewed by the authors before use.
We take responsibility for the final content of this work, including text, claims or artifacts produced
with the aid of generative AI.

\subsection*{Ethics statement}

This work uses existing 3D scenes to study text-controlled physical simulation. The human evaluation involved 25 participants who ranked simulated motions by their alignment with text instructions. Method identities were hidden as A, B, and C, their presentation positions were balanced across instructions, and results are reported in aggregate. Because the motion plans and automated evaluations rely on pretrained multimodal models, they may reflect model errors or biases. We therefore report human judgments alongside model-based scores and do not treat the latter as a sole measure of physical correctness.

\subsection*{Reproducibility statement}

Sections 3.1 and 3.2 describe the motion-planning and physics-optimization procedure and define the Motion-IG metric. Section 4.1 identifies the scene sources, text instructions, baselines, and evaluation models. The appendix provides additional results (Appendices A and B), examples of semantic injection (Appendix C), the prompts used for motion planning, code generation, and evaluation (Appendix D), and the human evaluation procedure (Appendix E).

\bibliography{iclr2027_agentdreamer}
\bibliographystyle{iclr2027_conference}

\clearpage
\appendix
% !TEX root = ../main_agentdreamer.tex

\section{Additional Qualitative Results}
\begin{figure}[h]
    \centering
    \includegraphics[width=\linewidth]{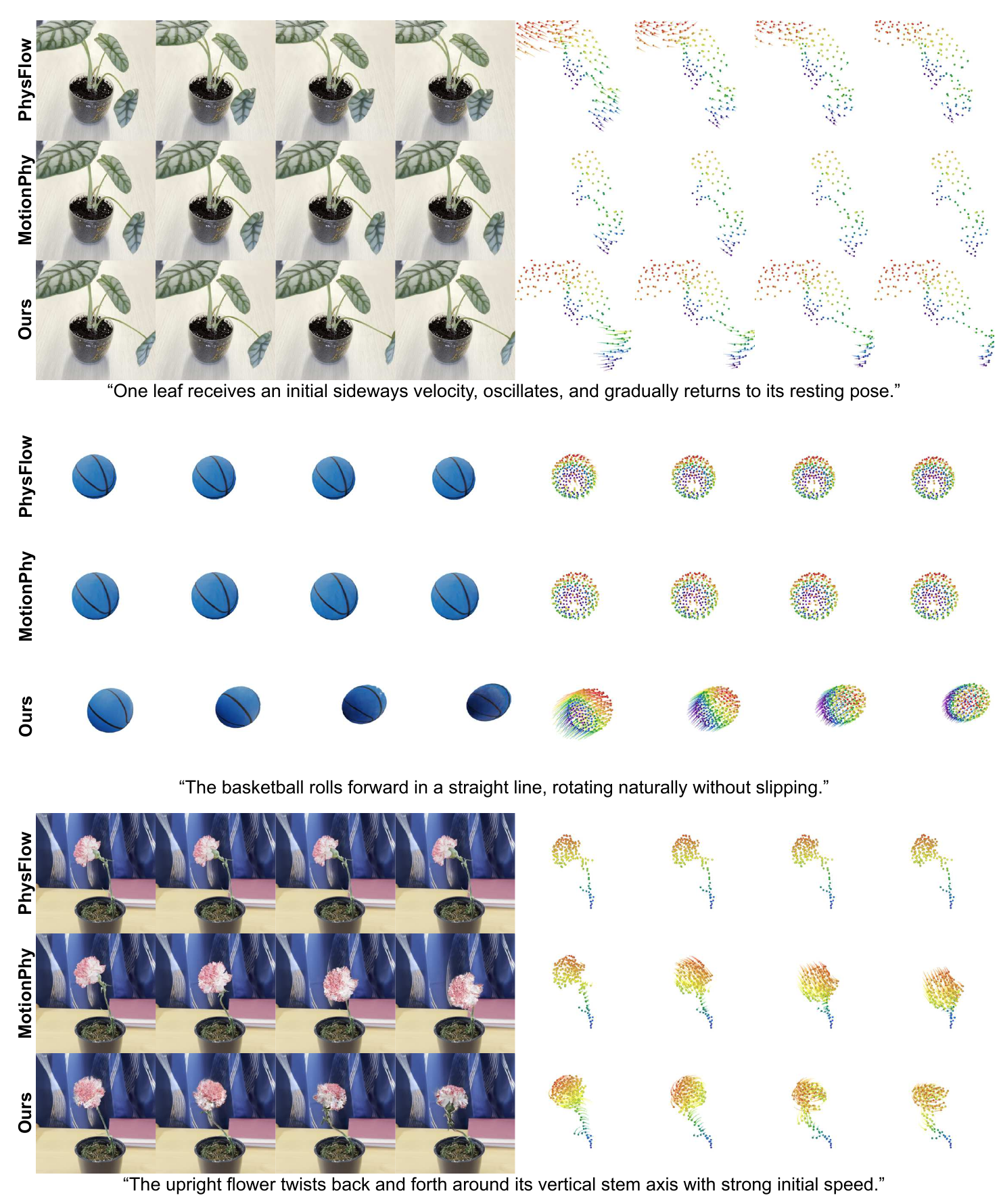}
    \caption{
        Additional qualitative comparisons with PhysFlow~\citep{physflow} and MotionPhysics~\citep{wang2026motionphysics} under the same text instructions. 
        The left panels show frames from the simulated videos, while the right panels show the corresponding sampled-point trajectories.
        In the shown examples, \method{} produces simulated motion that more closely follows the requested behavior.
    }
    \label{fig:supple_qualitative_results}
\end{figure}

\begin{figure}[h]
    \centering
    \includegraphics[width=\linewidth]{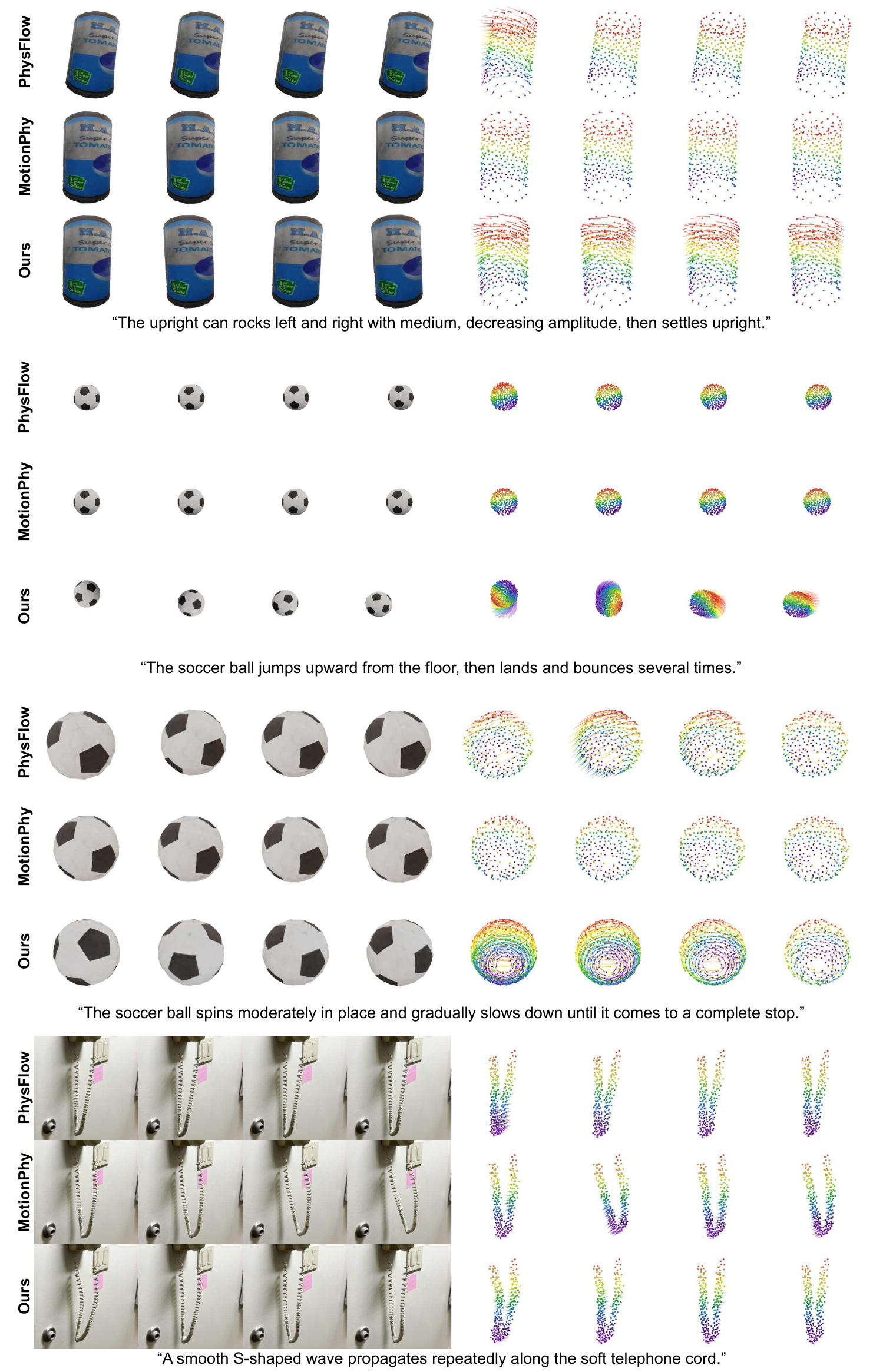}
    \caption{
        Additional qualitative comparisons with PhysFlow~\citep{physflow} and MotionPhysics~\citep{wang2026motionphysics} under the same text instructions.
        The left panels show frames from the simulated videos, while the right panels show the corresponding sampled-point trajectories.
    }
    \label{fig:supple_qualitative_results_2}
\end{figure}

Figure~\ref{fig:supple_qualitative_results} and Figure~\ref{fig:supple_qualitative_results_2} present additional comparisons of PhysFlow, MotionPhysics and \method{} under the same text instructions.
In the shown examples, \method{} produces motion that more closely follows the instructions than the baselines.
These results complement the main-paper comparisons and further support the use of agent-generated 3D motion priors for text-controllable physics simulation.

% \clearpage
% \section{Analysis of Parametric Motion Planning}

% \paragraph{Correlation with Known-Physics Dynamics? Equation?}

% - real physics

% - 

% - 
% - 

% \paragraph{smoothstep}

% \subsection{}

% \subsection{}

\clearpage

\section{Additional Quantitative Results}

\paragraph{Text--motion alignment across motion categories}
Table~\ref{table:category_alignment_supp} breaks the \ourmetric{} scores down by the motion categories of the main paper.
\method{} ranks first in every category under both open-weight and black-box evaluators.
The gap is clear for simple rigid-body motions governed by a single physical law, namely rolling, falling and bouncing, and spin decay: on rolling, \method{} reaches $+0.46$ with open-weight evaluators while both baselines stay negative ($-0.22$ and $-0.24$).
This is where the benefit of planning shows most, since the plan can state the motion by its governing law.
On more complex motions that repeat or change shape, oscillation and bending and deformation, \method{} still scores highest but by a narrower margin; with black-box evaluators it reaches $+0.35$ on oscillation, close to PhysFlow at $+0.33$.

\begin{table}[h]
\centering
\caption{
Text--motion alignment per motion category measured by \ourmetric{} (higher is better), averaged over four open-weight and five black-box evaluators.
}
\label{table:category_alignment_supp}
\vspace{-1.5mm}
\scalebox{0.78}{%
    \setlength{\tabcolsep}{10pt}
    \begin{tabular}{l|c|ccc|ccc}
        \toprule
        & & \multicolumn{3}{c|}{\textbf{Open-weight (avg.)}}
        & \multicolumn{3}{c}{\textbf{Black-box (avg.)}} \\
        \cmidrule(lr){3-5}\cmidrule(lr){6-8}
        \textbf{Category} & $\boldsymbol{n}$
        & \textbf{PhysFlow} & \textbf{MotionPhysics} & \textbf{Ours}
        & \textbf{PhysFlow} & \textbf{MotionPhysics} & \textbf{Ours} \\
        \midrule
        Fall and bounce & 4
        & $+0.18$ & $+0.36$ & $\mathbf{+0.63}$
        & $-1.46$ & $-0.87$ & $\mathbf{+0.77}$ \\
        Roll & 3
        & $-0.22$ & $-0.24$ & $\mathbf{+0.46}$
        & $-1.23$ & $-2.01$ & $\mathbf{+0.92}$ \\
        Spin decay & 2
        & $+0.29$ & $+0.20$ & $\mathbf{+0.55}$
        & $-0.15$ & $-0.23$ & $\mathbf{+0.62}$ \\
        Oscillation & 9
        & $+0.77$ & $+0.75$ & $\mathbf{+0.83}$
        & $+0.33$ & $-0.50$ & $\mathbf{+0.35}$ \\
        Bend and deform & 5
        & $+0.67$ & $+0.70$ & $\mathbf{+0.74}$
        & $+0.34$ & $-0.12$ & $\mathbf{+0.39}$ \\
        \bottomrule
    \end{tabular}
}
\vspace{-2mm}
\end{table}

\paragraph{Human evaluation details}
We collect rankings from 25 participants on 20 instructions, for 500 trials in total.
In each trial, participants read the instruction and watch the results of the three methods, stacked vertically and labeled A, B, and C without method names.
The position of each method is balanced across instructions.
Participants rank the three results from first to third by how well the motion matches the instruction.
Table~\ref{table:human_rank_supp} reports how often each method receives each rank.
\method{} is ranked first in 73.6\% of trials and last in only 5.4\%, whereas PhysFlow and MotionPhysics are ranked last in 40.2\% and 54.4\% of trials, respectively.

\begin{table}[h]
\centering
\caption{
Rank distribution in the human evaluation (25 participants $\times$ 20 instructions $=$ 500 trials).
Each entry is the percentage of trials in which the method receives the rank.
}
\label{table:human_rank_supp}
\vspace{-1.5mm}
\scalebox{0.78}{%
    \setlength{\tabcolsep}{14pt}
    \begin{tabular}{l|ccc}
        \toprule
        \textbf{Method} & \textbf{1st (\%)} $\uparrow$ & \textbf{2nd (\%)} & \textbf{3rd (\%)} $\downarrow$ \\
        \midrule
        PhysFlow      & 14.0 & 45.8 & 40.2 \\
        MotionPhysics & 12.4 & 33.2 & 54.4 \\
        Ours          & \textbf{73.6} & 21.0 & \textbf{5.4} \\
        \bottomrule
    \end{tabular}
}
\vspace{-2mm}
\end{table}

\clearpage

\section{Semantic Injection Examples}

\begin{figure}[h]
    \centering
    \includegraphics[width=\linewidth]{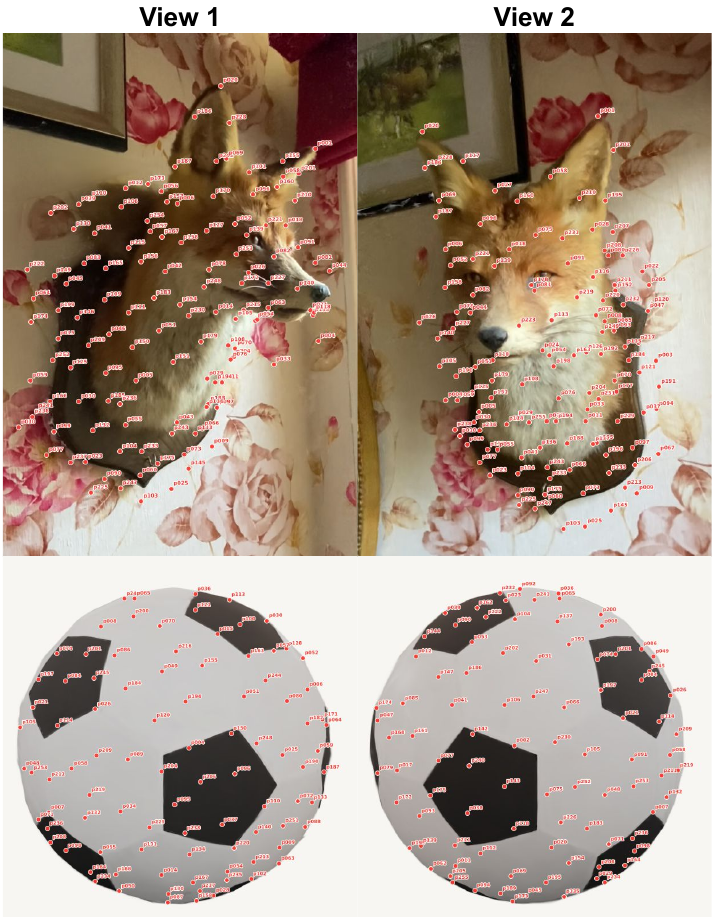}
    \caption{
        Examples of annotated images used for automatic semantic injection. Control points sampled from the initial scene geometry are marked with numerical identifiers in two views of a fox (top) and a soccer ball (bottom).
    }
    \label{fig:supple_semantic_injection}
\end{figure}

Figure~\ref{fig:supple_semantic_injection} presents examples of annotated images for two objects. The numerical identifiers show where the sampled points appear in each initial scene image, helping the agent associate the points with visible object parts relevant to the text instruction.

\clearpage

\section{Details of Agent Prompts}

\subsection{Prompts for Motion planning}
\begin{figure}[h]
    \centering
    \includegraphics[width=0.4\linewidth]{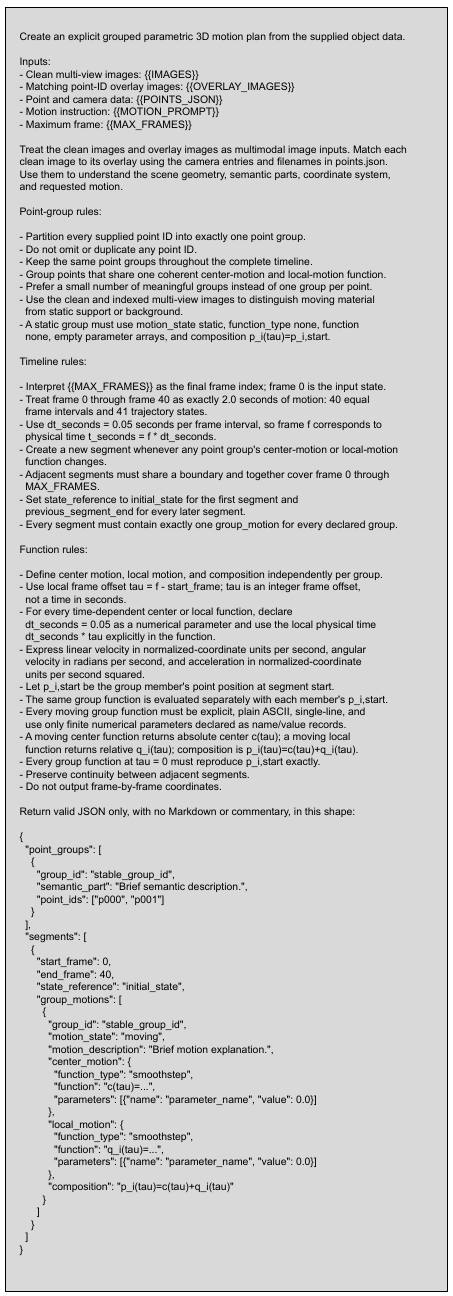}
    \caption{
        Prompt for agent-based motion planning. 
        The agent divides the time horizon into temporal chunks and specifies parametric motion functions for groups of sampled points.
    }
    \label{fig:supple_plan_prompt_1}
\end{figure}

\begin{figure}[h]
    \centering
    \includegraphics[width=0.8\linewidth]{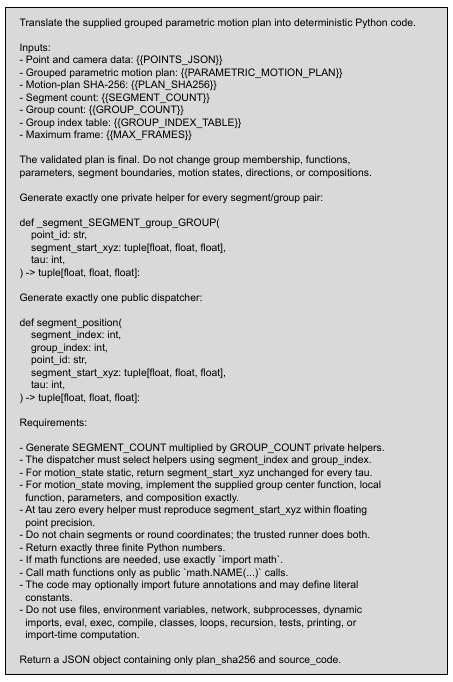}
    \caption{
        Prompt for translating the motion plan into deterministic Python code. 
        The code implements the specified motion functions to compute target point positions.
    }
    \label{fig:supple_plan_prompt_2}
\end{figure}

Figure~\ref{fig:supple_plan_prompt_1} presents the prompt used to construct the motion plan. The agent groups the sampled points, divides the time horizon into temporal chunks, and specifies a center function and a local function for each group in each chunk. Figure~\ref{fig:supple_plan_prompt_2} presents the prompt used to translate the resulting motion plan into deterministic Python code. The code computes target point positions from the specified functions without modifying the motion plan.

\clearpage

\subsection{Prompts for Evaluation}
\begin{figure}[h]
    \centering
    \includegraphics[width=0.6\linewidth]{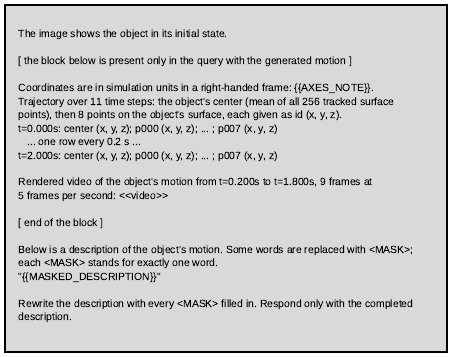}
    \caption{
        Prompt for the open-weight evaluators.
        The masked words are scored by teacher forcing, and the same prompt without the motion gives the reference term.
    }
    \label{fig:supple_eval_prompt_1}
\end{figure}

\begin{figure}[h]
    \centering
    \includegraphics[width=0.6\linewidth]{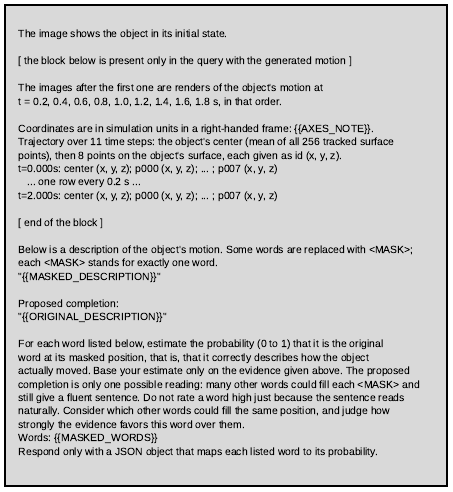}
    \caption{
        Prompt for the black-box evaluators.
        Each masked word receives a stated probability, collected in one JSON response per query.
    }
    \label{fig:supple_eval_prompt_2}
\end{figure}

Figure~\ref{fig:supple_eval_prompt_1} presents the prompt given to the open-weight evaluators. The evaluator receives the annotated initial views, the generated motion as a trajectory and as rendered frames, and the description with its motion words masked; the masked words are scored by teacher forcing rather than by sampling the completion. Figure~\ref{fig:supple_eval_prompt_2} presents the prompt given to the black-box evaluators. Since current API models expose no teacher-forced log-probabilities, the evaluator states a probability for each masked word, and these ratings replace the log-probabilities in the same gain.

\clearpage
\section{Details of User Study}
We conducted a user study to assess how closely the simulated motions follow their text instructions. 
For each of 20 instructions, participants viewed three videos of the same object, corresponding to PhysFlow, MotionPhysics, and \method{}, and ranked them from first to third according to their text–motion alignment. 
At the beginning of the survey, we present the following instructions:

\begin{quote}
\itshape
In each of the 20 questions, you will see three short videos (A, B, C) of the same object and a text description below them. The videos loop automatically. For each question, please rank the three videos 1st, 2nd, and 3rd: 1st is the one whose motion best matches the text, and 3rd is the one that matches it the least. Each rank can be used only once.

Please judge only how well the motion follows the description: what moves, in which way, and how strongly. Do not judge image quality, lighting, or rendering artifacts. The key words in each description are highlighted in orange. Please watch all three videos at least once before answering.
\end{quote}

\begin{figure}[h]
    \centering
    \includegraphics[width=0.4\linewidth]{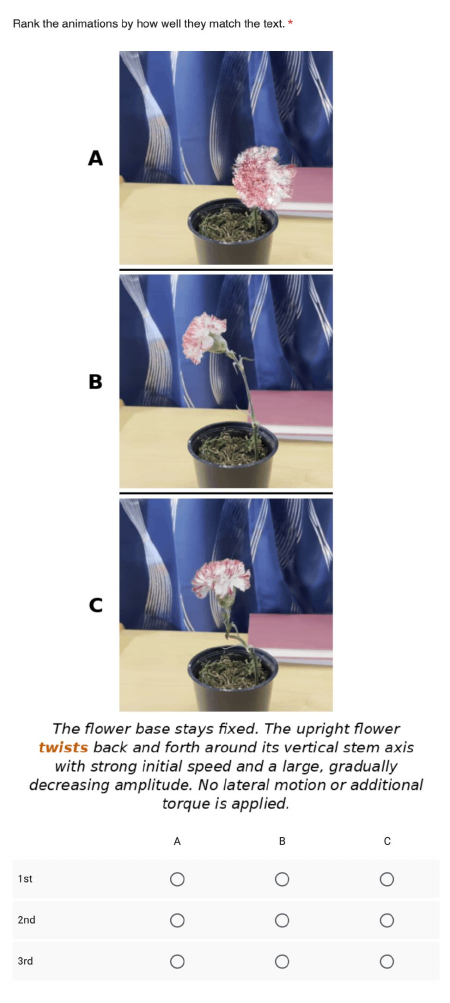}
    \caption{
        Example question in the human evaluation of text--motion alignment. 
        Given a text instruction and three simulation results for the same object, participants rank the motions by how closely they follow the requested behavior.
    }
    \label{fig:supple_user_study}
\end{figure}

Figure~\ref{fig:supple_user_study} shows an example question presented to participants in the user study.

\clearpage

\end{document}